# Abjad-Kids: An Arabic Speech Classification Dataset for Primary Education

Abdul Aziz Snoubara, Baraa Al_Maradni, Haya Al_Naal, Malek Al_Madrmani, Roaa Jdini, Seedra Zarzour, Khloud Al Jallad*

Arab International University, Daraa, Syria.
* Corresponding Author: k-aljallad@aiu.edu.sy

## Abstract

Speech-based AI educational applications have gained significant interest in recent years, particularly for children. However, children speech research remains limited due to the lack of publicly available datasets, especially for low-resource languages such as Arabic. This paper presents Abjad-Kids, an Arabic speech dataset designed for kindergarten and primary education, focusing on fundamental learning of alphabets, numbers, and colors. The dataset consists of 46,397 audio samples collected from children aged 3–12 years, covering 141 classes. All samples were recorded under controlled specifications to ensure consistency in duration, sampling rate, and format. To address high intra-class similarity among Arabic phonemes and the limited samples per class, we propose a hierarchical audio classification based on CNN-LSTM architectures. Our proposed methodology decomposes alphabet recognition into a two-stage process: an initial group-splitting model followed by specialized classifiers for each group. Both strategies: static linguistic-based grouping and dynamic clustering-based grouping, were evaluated. Experimental results demonstrate that static linguistic-based grouping achieves superior performance. Comparisons between traditional machine learning with deep learning approaches, highlight the effectiveness of CNN-LSTM models combined with data augmentation. Despite achieving promising results, most of our experiments indicate a challenge with overfitting, which is likely due to the limited number of samples, even after data augmentation and model regularization. Thus, future work may focus on collecting additional data to address this issue. Abjad-Kids will be publicly available. We hope that Abjad-Kids enrich the children's representation in speech dataset, and be a good resource for future research in Arabic speech classification for kids.

## 1- Introduction

While adult speech classification has matured significantly through decades of rigorous research, these models often fail to generalize to younger demographics due to the unique acoustic and linguistic characteristics of child speech. Consequently, the child speech classification domain remains comparatively under-explored [1] [2] [3]. Despite its vital importance in educational technology, specifically in Computer-Assisted Language Learning

(CALL), progress in this area is currently bottlenecked by a lack of diverse, high-quality datasets. This scarcity persists even though such research is essential for effectively teaching languages to children.

Teaching languages to children is a challenging task, often become harder by the lack of truly interactive tools. Fortunately, AI is beginning to change the landscape, offering a way to turn language learning into something more than just rote memorization. By creating engaging, personalized learning environments, we can help children stay motivated and connected to their studies [4]. When these smart tools are built on solid educational theories like constructivism, they allow children to engage in active learning by mimicking social dialogue and negotiating meaning, which significantly improves retention [5]. This approach makes the whole experience much more effective by allowing children to learn at their own pace. To be truly successful, the design of these educational tools needs to be intuitive, focusing on interactivity, self-learning, and the ability to adapt to a child's specific level. Integrating visual and auditory activities helps children retain new information much more effectively. This is where voice-driven technology becomes a game-changer. By incorporating speech-based interaction into smart toys and educational platforms which provides real-time feedback and aids in pronunciation, we allow children to move away from rigid keyboards and touchscreens and interact with language naturally. These adaptive environments do more than just teach; they make the process feel personal, significantly helping to maintain motivation and reduce the likelihood of children dropping out of their language programs.

Speech-based interaction has become a fundamental component of modern educational technologies, particularly in applications designed for early childhood learning. Despite the significant progress achieved in speech recognition and audio classification for high-resource languages such as English, Arabic remains a relatively underexplored language in this domain, particularly for children's speech, due to several factors:

- **Linguistic Complexity:** Arabic possesses a rich phonetic inventory with subtle articulatory differences between letters [6].
- **Acoustic Variability:** Children's speech is characterized by inconsistent pronunciation and developmental variability, making models trained on adult speech ineffective [7] [8].
- **Data Scarcity:** There is a significant lack of publicly available, large-scale speech datasets specifically for children's speech [1] [2], especially in Arabic.

To address these challenges, we proposed a hierarchical classification methodology. Our proposed methodology is a two-stage strategy for decomposing complex, overwhelming classification task into simpler sub-tasks. By first grouping similar classes based on linguistics articulation points and then applying specialized classifiers within each group. Hierarchical approaches can reduce class confusion and improve overall performance even with high intra-class similarity of Arabic letters due to similar pronunciation patterns, especially when pronounced by children. We believe that this methodology is particularly well-suited for Arabic speech, where linguistic knowledge such as articulation points and phonetic properties can be leveraged to construct meaningful class groupings.

Another major obstacle in Arabic speech classification researches is the lack of publicly available, large-scale, and well-annotated datasets, especially for children. Most existing Arabic speech datasets focus on adult speakers and are primarily designed for automatic speech recognition rather than keyword spotting or speech classification. Moreover, the few available Arabic children speech datasets are typically limited in size, lack diversity in recording conditions, or do not provide sufficient class coverage for educational applications. This scarcity of data significantly restricts the development, evaluation, and comparison of robust keyword spotting models for Arabic children speech.

The main contributions of this paper are:
- A new Arabic speech classification benchmark, Abjad-Kids dataset that contains 46,397 samples from children aged 3–12, covering alphabets, numbers, and colors.
- A two-stage hierarchical methodology for Arabic alphabets classification. First stage is grouping classification model based on linguistic articulation points followed by the second stage that contains specialized classifier for each linguistic group.

The remainder of this paper is organized as follows: section 1 is an introduction; Section 2 discusses related works, and Section 3 describes the dataset construction methodology. Section 4 presents our evaluation experiments. Results and discussions are shown in Section 5. Finally, conclusion, future works and limitations are shown in sections 6, 7, respectively.

**2- Related Works**
In this section, we will discuss state of the art children speech datasets and models.

**2.1. Children Speech Datasets**

While the landscape of children speech resources remains specialized, several seminal datasets have historically defined the field's development. The following list summarizes key corpora:

- **TIDIGITS** [9]**:** A foundational corpus created in 1982**,** comprising 13 hours of English digit sequences spoken by over 300 men, women, and children. It was built for the purpose of designing and evaluating algorithms for speaker-independent recognition of connected digit sequences. It features 326 speakers, including 101 children (50 boys and 51 girls) each pronouncing 77 digit sequences.
- **CMU Kids** [10]: It is designed for the LISTEN reading coach project at Carnegie Mellon University in 1996. It contains 5,180 utterances from children aged 6–11. It is split into **SUM95** that focused on "good" readers and contained 44 speakers and 3,333 utterances in this set, and **FP** that focused on errorful reading and dialectal variants like "Pittsburghese" and contained 32 speakers and 1,847 utterances in this set, providing crucial data for educational technology.
- **OGI Kids** [12]**:** The corpus was created in 2000 and it is composed of both prompted and spontaneous speech from 1100 children from kindergarten through grade 10. The prompted speech was presented as text appearing below an animated character (Baldi) that produced accurate visible speech synchronized with recorded prompts. The speech and text consist of isolated words, sentences, and digit strings.
- **SEED** [11]: A specialized database providing examples of adult and child speech production, covering both typically developing individuals and those with speech disorders, collected under highly controlled conditions.
- **SLT 2021 CSRC** [3] A children speech recognition challenge dataset offering 400 hours of Mandarin speech, designed in 2021 to benchmark children Automatic Speech Recognition performance in high-variability environments.
- **MyST** [13]:**:** it was developed as part of the My Science Tutor project in 2023. It is One of the largest collections of conversational speech, comprising 400 hours and 230,000 utterances across about 10.5K virtual tutor sessions by around 1.3K third, fourth and fifth grade students.
- **Arabic Little STT** [6]: a dataset of Levantine Arabic child speech, containing 355 utterances from 288 children (ages 6 - 13) and recorded in classroom environments.
- **ChildMandarin (2025)** [14]: Mandarin speech dataset created in 2025. It comprises 41.25 hours of speech with care- fully crafted manual transcriptions, collected from 397 speakers across various provinces in China, with balanced gender representation.

To mitigate the scarcity of high-quality children speech data, researchers have increasingly turned to synthetic data augmentation. For instance, [15] address the inherent variability of children's speech and the limitations of training corpora by employing Text-to-Speech (TTS) systems to generate synthetic training samples.

## 2.2. SoTA Models

The methodology for evaluating children's speech has moved through three distinct technological eras, each solving the acoustic disparity problem between children and adults differently. Early SOTA research relied on Hybrid Deep Neural Network as noted by [8], the primary challenge was the acoustic mismatch, where transfer learning was employed to adapt adult-trained models to children's voices, yet these models often struggled with non-linear physiological differences like shorter vocal tract lengths and higher fundamental frequencies. The field subsequently moved toward Transformer-based Self-Supervised Learning (SSL) using foundation models such as Whisper, Wav2vec 2.0, HuBERT, and WavLM [16] which utilize Transformer backbones to capture long-range temporal dependencies; however, because these models are pre-trained on adult speech, they suffer from domain shift as they are biased to the pretraining data. When SSL models are finetuned with data from another domain, domain shifting occurs and might cause limited knowledge transfer for downstream tasks. Domain Responsible Adaptation and Fine-tuning framework (DRAFT) [17] was proposed to solve this gap. DRAFT uses both causal and non-causal transformers to bridge the domain gap and yield up to 30% relative improvements in Word Error Rate (WER). Most recently, research has prioritized model efficiency and data resilience, with Kid-Whisper[2] demonstrating that refined preprocessing can extremely reduce WER, while other researchers [15] employ Adapter-based architectures with Text-to-Speech (TTS) augmentation; by training lightweight bottleneck adapters on synthetic data while keeping the main pre-trained model frozen, they achieved a 6% relative reduction in WER, proving that models can learn effectively even from imperfect synthetic pediatric audio. As for [18], this study aimed to develop a classifier to differentiate child and adult speech using datasets such as LibriSpeech, VoxCeleb, Common Voice, OGI Kids, CMU Kids, and MyST. Initial attempts using diarization and embeddings from ECAPA-TDNN, Whisper, and HuBERT led to misclassification due to dataset-specific clustering caused by recording artifacts, they highlight that model can still overfit to recording artifacts rather than age-based features, confirming that while impulse response augmentation and fine-tuned classifiers are current best practices, the scarcity of diverse, high-volume datasets in regional languages remains a critical bottleneck for future model generalization.

### 3- Dataset Construction Methodology

This section details the methodology for constructing the Abjad-Kids dataset, which was designed to bridge the existing gap in Arabic children's speech resources, specifically for keyword spotting tasks in early education.

#### 3.1. Dataset Collection

Collecting high-quality audio data from kids presents significant logistical and practical challenges. To facilitate large-scale data collection and ensure consistency across recordings, this project was conducted by collaborating with multiple primary schools and kindergartens in Syria. Moreover, a dedicated desktop application was developed to streamline the recording process and facilitate data organization. The application featured a child-friendly interface designed to keep participants engaged and minimize recording errors. Furthermore, this data collection application supported the simultaneous recording of two children, which promoted peer learning and helped create a more comfortable environment. To ensure high-quality and standardization, all recordings were captured using identical hardware (KM headphones) and stored according to the following specifications:

- **Duration:** 2 seconds
- **Sample Rate:** 16 kHz
- **Channels:** Mono (1 channel)
- **Format:** WAV

### 3.2. Dataset Analysis

The final dataset contains 46,397 audio samples collected from children aged 3–12 years. These include 36,751 alphabet samples, 6,649 number samples, and 2,997 color samples, corresponding to a total duration of approximately 25.77 hours of audio. The data collection process required approximately 66 hours of recording sessions conducted across multiple primary schools and kindergartens in Syria. The dataset consists of 141 distinct classes distributed as follows: 112 alphabet-related classes, 20 number classes, and 9 color classes. The alphabet classes include the 28 Arabic letters and three commonly used words associated with each letter, resulting in a diverse vocabulary suitable for educational applications. Number classes cover digits from 0 to 9 as well as multiples of ten up to 100, while color classes include the most commonly taught colors for early learners. The distribution of samples across the target categories is summarized in Table 1. While the dataset represents a diverse and one of the largest Arabic children's speech datasets for keyword spotting currently available, we observed class imbalances. Consequently, data augmentation techniques were employed during the model training phase to improve robustness.

| Category | # Classes | # Samples | Approx. Duration |
|---|---|---|---|
| **Alphabets** | 112 | 36,751 | ~20 hours |
| **Numbers** | 20 | 6,649 | ~4 hours |
| **Colors** | 9 | 2,997 | ~1.7 hours |
| **Total** | **141** | **46,397** | **25.77 ours** |

*Table 1: Abjad-Kids Dataset Samples Distribution*

## 4. Evaluation Experiments

To evaluate our constructed dataset, Abjad-Kids, we implemented several models for handling the audio classification tasks. The fundamental concept is to create a model that receives an audio sample as input and outputs the corresponding class.

The audio dataset will be divided into three sections: alphabets, numbers, and colors. However, the first step before any model training is a preprocessing stage applied to the audio samples.

### 4.1. Audio Preprocessing:

Recording audio samples in primary schools and kindergartens is challenging due to persistent background noise, which poses a significant difficulty for data quality. Thus, our proposed pipeline first preprocess data to ensures that it is clean and consistent before it is fed into the classification models. As illustrated in Figure 1, the preprocessing stages are as follows:

- **Resampling Audio**: To ensure standardization and consistency, all audio samples are standardized to a sampling rate of 16 kHz. While the initial collection targeted 16 kHz, this step verifies that every file maintains a consistent frequency domain, which is a prerequisite for model stability.

- **Noise Filtering**: Given that background noise is a persistent challenge in primary schools and kindergartens, effective noise reduction is crucial. To mitigate this interference, a masking technique was applied to isolate the signal from background noise. This mask identifies and preserves portions of the signal that exceed a defined noise power threshold.

- **Pre-Emphasis**: To enhance the high-frequency components of the audio, a pre-emphasis technique was employed. This process involves applying a high-pass filter that amplifies the higher frequencies, which helps balance the spectral energy and improves the feature extraction process.

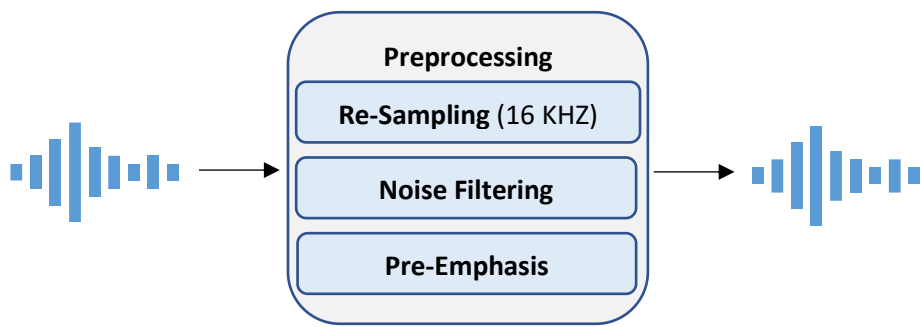

*Figure 1: Preprocessing Applied in All Experiments*

## 4.2. Data Augmentation

Due to the limited number of samples per class, data augmentation was employed to improve model generalization and reduce overfitting. As shown in Figure 2, each audio sample was augmented multiple times using pitch shifting, low-pass filtering, and gain adjustment. These transformations simulate realistic variations in pitch, frequency content, and loudness while preserving the semantic meaning of the keyword.

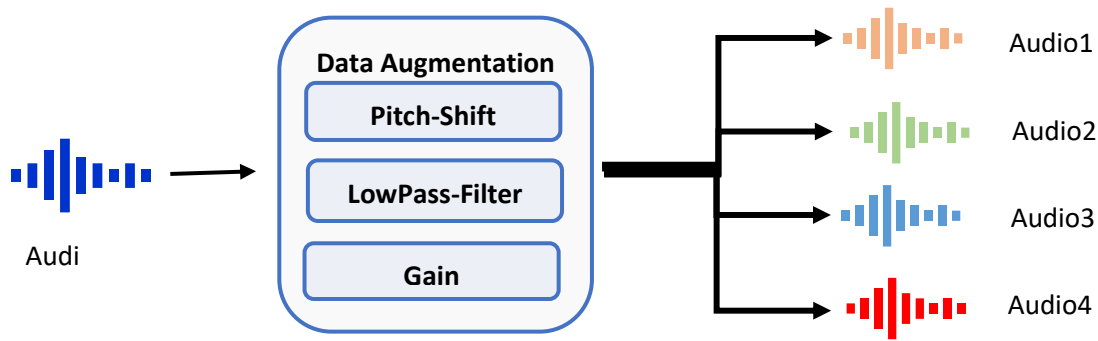

*Figure 2: Data Augmentation*

## 4.3. Feature Extraction

Several features were extracted and evaluated to determine the most effective set or combination of sets for keyword spotting, the extracted features are as follows:

- **Time domain features:**
  - Zero Crossing Rate (ZCR) counts the number of times the audio signal crosses the zero-amplitude line within a given time frame, often correlated with higher frequencies.
  - Short Term Energy (STE) measures the signal's power within a short time frame, reflecting the amplitude or intensity of the sound.
- **Frequency domain features:**
  - Spectral Centroid (SC) represents the center of mass of the spectrum, indicating the dominant frequency.
  - Spectral Entropy (SE) measures the randomness in the spectral distribution.
  - Spectral Flux (SF) quantifies the change in the spectrum between successive frames, revealing how much the frequency content is evolving.
  - Spectral Roll-off (SR) indicates the frequency below which a specified percentage of the total spectral energy is contained.
  - MFCC (13): a set of 13 features capturing the short-term power spectrum of a sound, based on human perception.

To evaluate the impact of feature representation on model performance, two extraction techniques were compared:

- **Mean Aggregation**: Extracts only the mean value for each feature, resulting in a 19-dimensional feature vector.
- **Statistical Aggregation**: Extracts the mean, standard deviation, skewness, max, min, and median for each feature, resulting in a 210-dimensional feature vector.

### 4.4. Class Grouping Strategies

Direct classification of all alphabet-related classes poses a significant challenge due to the high number of classes and the limited number of samples per class. With 112 classes (28 characters and 84 words) and only approximately 310 samples per class, the data density per category is relatively low. To address this issue, a hierarchical classification strategy was employed, where the alphabet recognition task is decomposed into two stages: class grouping followed by group-specific classification. Two grouping strategies were investigated (shown in Figure 7): static grouping and dynamic grouping.

#### 4.4.1. Static Grouping

The static grouping strategy is based on linguistic knowledge of Arabic phonetics and places of articulation. Arabic characters are categorized according to their specific places of articulation, such as the throat (halq), tongue (lisan), and lips (shafatan), which group characters with similar pronunciation mechanisms. We grouped all classes into six distinct groups based on these phonetic regions. The Nasal (Al-Khayshum) category was excluded as all its characters are represented within other groups, while the Upper/Lower Incisors were included as distinct categories.

By leveraging well-established Arabic pronunciation rules, the static grouping strategy produces semantically and phonetically meaningful groups. This approach reduces inter-group confusion and simplifies the classification task. Being independent of the training data distribution, this method is particularly robust for low-resource scenarios.

The detailed Arabic alphabet phonetic places of articulation are shown in Figure 3, and the resulting static groups are illustrated in Figure 4.

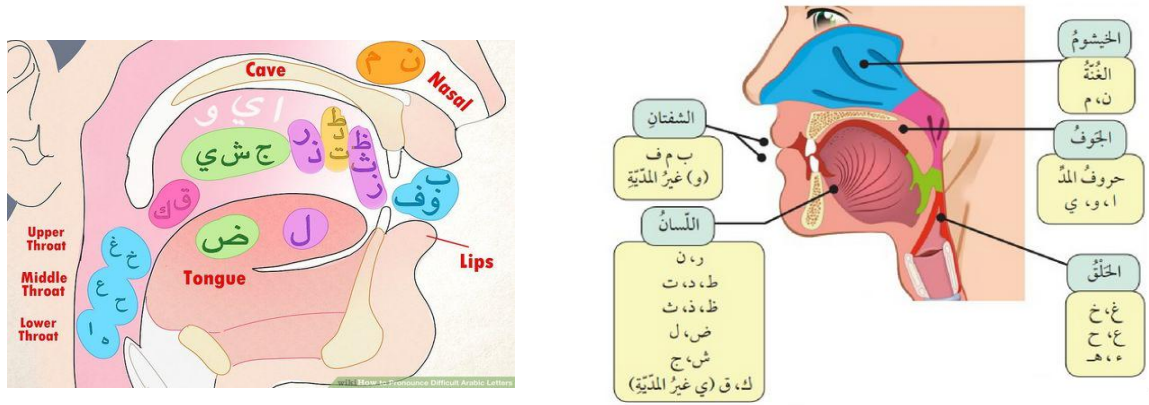

*Figure 3: Arabic Alphabet Phonetic Place of Articulation [1,2]*

### 4.4.2. Dynamic Grouping

In contrast, the dynamic grouping strategy employs data-driven clustering techniques to automatically group similar classes. This approach uses feature representations extracted from audio samples as the input for clustering algorithms. Initially, the Elbow method was used to determine the optimal number of clusters. The results indicated that nine clusters provided the most suitable grouping as shown in Figure 5. Subsequently, the K-means algorithm was utilized to assign classes to these clusters based on feature similarity. To further analyze inter-cluster relationships, hierarchical clustering was applied to the resulting centroids, providing a visual representation of how these clusters relate to one another. The resulting cluster assignments are illustrated in Figure 6.

While dynamic grouping offers flexibility and adaptability to data characteristics, its effectiveness is highly dependent on the quality and quantity of available samples.

---

[1] https://shneler.com/2020/7/is7/f1/xxxxx/
[2] https://surahquran.com/Tajweed/letter-exits.html

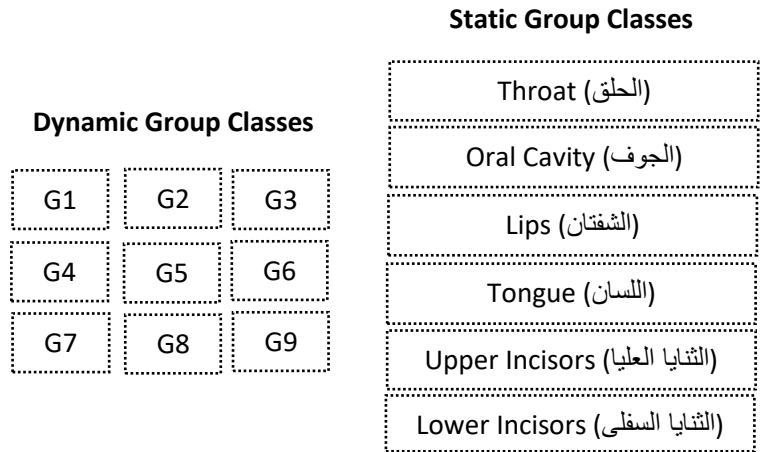

Figure 4: Dynamic Grouping Classes vs. Static Grouping Classes

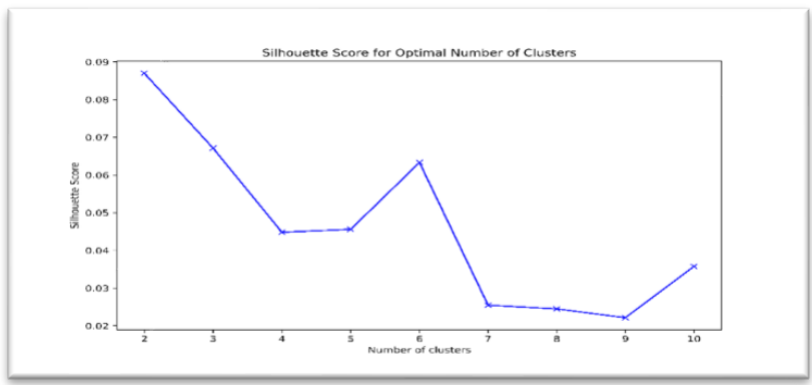

Figure 6: Elbow for Finding Optimal Number of Clusters

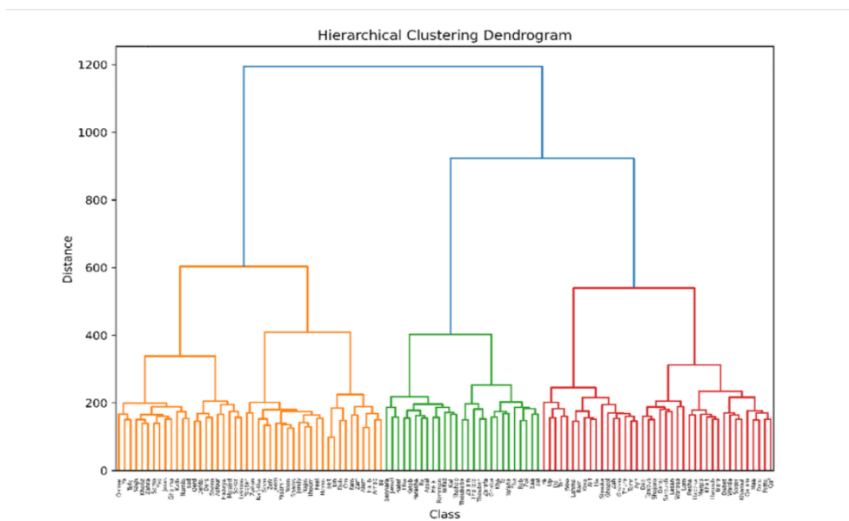

Figure 5: Hierarchical Clustering Dendrogram

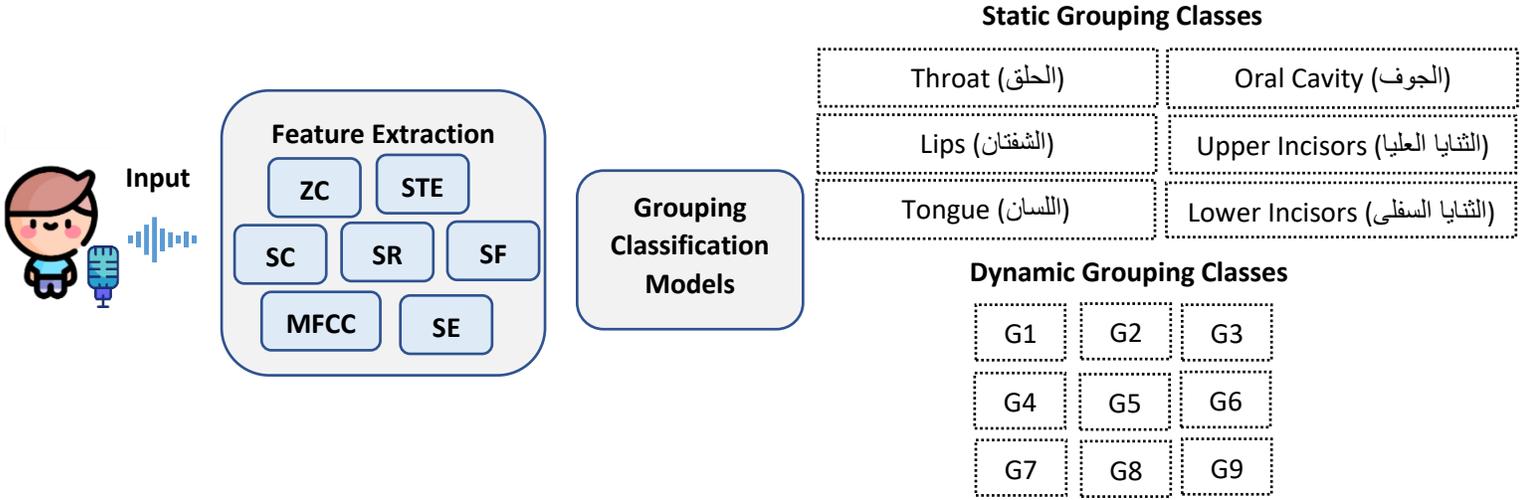

*Figure 7: Main Grouping Classification Proposed Methodology*

### 4.4.3. Proposed Architecture

Recent advances in deep learning have significantly improved performance in audio classification tasks. Convolutional Neural Networks (CNNs) have demonstrated strong capabilities in extracting local time–frequency patterns from audio representations such as Mel-frequency cepstral coefficients (MFCCs), while Recurrent Neural Networks (RNNs) and Long Short-Term Memory (LSTM) networks are effective in modeling temporal dependencies in speech signals. Hybrid CNN-LSTM architectures combine the strengths of both approaches and have shown promising results in keyword spotting and speech recognition tasks. Nevertheless, training deep learning models typically requires large amounts of labeled data, which remains a critical limitation in Arabic children speech applications. Thus, the proposed deep learning architecture combines Convolutional Neural Networks (CNNs) and Long Short-Term Memory (LSTM) networks. CNN layers are responsible for extracting local time–frequency patterns from MFCC representations, while LSTM layers model temporal dependencies across frames. Fully connected layers are then used for classification. Multiple CNN-LSTM architectures were evaluated by varying the number of convolutional layers, filter sizes, dense layers, dropout rates, and LSTM units (shown in Section 5). In addition to training models from scratch, Moreover, we compared our results with a baseline CNN-based architecture reported in previous works for keyword spotting in noisy environment [19].

### 4.5. Group-Specific Classification Models:

Following the initial class grouping phase, a dedicated model is trained for each specific category. This hierarchical architecture enables the system to first identify the broad phonetic group of an input, subsequently performing a more refined classification within that specific subset. Given that the static grouping strategy consistently outperformed the dynamic approach, all subsequent group-specific models were developed using the static configuration. The overall two-stage methodology is illustrated in Figure 8.

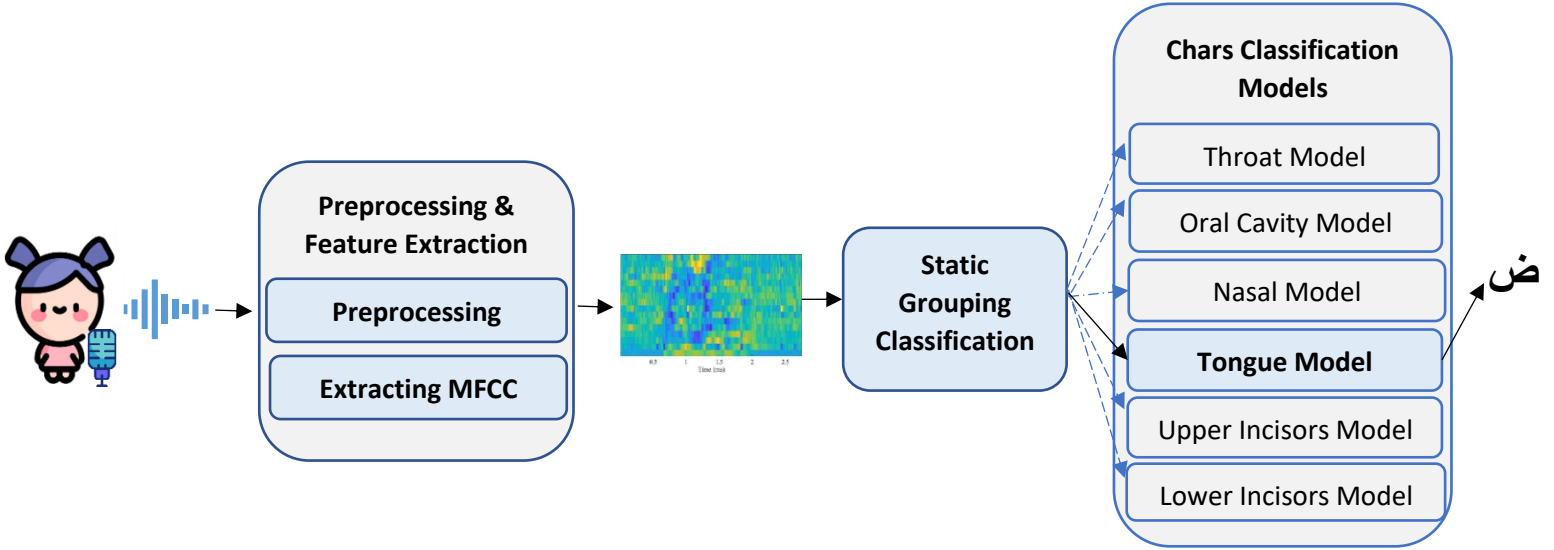

*Figure 8: Overview of the Proposed Two-Stage Methodology*

### 4.6. Experimental Setup

The dataset was divided into training and testing sets, ensuring that samples from the same speaker did not appear in both sets to reduce speaker-dependent bias. Model performance was evaluated using classification accuracy, which is a commonly used metric for isolated keyword spotting tasks. For deep learning models, training was performed using the Adam optimizer with varying learning rates depending on the experiment. Categorical cross-entropy was used as the loss function. Dropout and data augmentation were applied to mitigate overfitting.

## 5. Results and Discussion

This section presents evaluation methodology, and results obtained from both machine learning and deep learning models. The experiments aim to evaluate the effectiveness of the proposed hierarchical methodology, compare grouping strategies, and analyze specific-group models performance.

First, to evaluate the effectiveness of the proposed hierarchical classification strategies, we compared the performance of machine learning and deep learning static and dynamic grouping against a baseline architecture [20].

Given the poor performance of machine learning approaches, deep learning models were adopted for all subsequent experiments. CNN-LSTM architectures were evaluated for alphabet group splitting, alphabet group classification, number classification, and color classification.

As for machine learning experiments, machine learning models were evaluated using the handcrafted features described in Section 4.3 for grouping stage, which was challenging classification task due to the large number of classes and limited samples per class. Support Vector Machines (SVM), Random Forest (RF), and ensemble learning methods were evaluated. Grid search was employed to tune hyperparameters for each model. For SVM, different values of the regularization parameter $C$, kernel type, and gamma were tested. Random Forest models were evaluated with varying numbers of estimators, tree depth, and feature selection strategies. An ensemble model combining Logistic Regression, SVM, and Random Forest was also evaluated. The results indicate that machine learning models struggled to achieve satisfactory performance on the alphabet grouping task. The best-performing SVM model achieved an accuracy of 39%, while Random Forest and ensemble models achieved accuracies of 38%. These results highlight the limitations of handcrafted features and traditional classifiers in modeling the complex acoustic patterns of Arabic children speech.

As for grouping stage, Table 2 summarizes the training and test accuracy achieved across various model configurations. The results indicate that the Static Grouping strategy consistently outperforms both the dynamic approach and the baseline predefined architecture, achieving a maximum test accuracy of 78%. While the highest-performing static configuration achieved a training accuracy of 92%, the 14% generalization gap suggests that the model exhibits some degree of overfitting. This indicates that the 210-feature vector approach, while expressive, may capture noise in the limited training samples. In contrast, the configuration with an 81% training accuracy demonstrates a more stable generalization, with a narrower 7% gap, providing a more balanced performance. Overall, the static grouping method proves to be a robust approach for low-resource keyword spotting in early education, as it leverages domain-specific linguistic knowledge to reduce inter-class confusion.

| Grouping Strategy | Conv layers | Dens layers | Dropout | LSTM layers | Train accuracy | Test Accuracy |
|---|---|---|---|---|---|---|
| **Dynamic** | 3(128,64,32) | 3(256,32,16) | 0.3 | 2(64,32) | **72%** | **70%** |
| | 3(128,64,32) | 3(256,32,16) | 0.3 | 2(64,32) | 77% | 70% |
| | baseline architecture [20] | | | | 81% | 67% |
| **Static** | 3(64,64,32) | 1(128) | 0.25 | 2(64,32) | 81% | 74% |
| | 3(64,64,32) | 1(256) | 0.3 | 2(128,64) | **92%** | **78%** |
| | baseline architecture [20] | | | | 88% | 71% |

Table 2: Dynamic Grouping Results vs. Static Grouping Results

After static grouping, separate CNN-LSTM models were trained for each alphabet group. Extensive experiments were conducted (shown in Table 3) by varying the number of convolutional layers, dense layers, dropout rates, LSTM units, batch size, and learning rate.

| Group Name | Conv layers | Dense layers | Dropout | LSTM layers | Train accuracy | Test Accuracy |
|---|---|---|---|---|---|---|
| Aqsa-lessan | 3(128,64,32) | 2(256,32) | 0.15 | 2(128,32) | 95% | 89% |
|  | 3(64,64,32) | 2(256,32) | 0.25 | 2(128,32) | 93% | 88% |
|  | 3(128,64,32) | 2(256,16) | 0.4 | 2(64,32) | **93%** | **89%** |
| Halq | 3(128,64,32) | 3(512,256,32) | 0.25 | 2(128,128) | 91% | 84% |
|  | 3(128,64,32) | 3(256,64,32) | 0.3 | 2(64,32) | 91% | 83% |
|  | 3(128,64,32) | 3(512,64) | 0.1 | - | **99%** | **92%** |
| Jouf | 2(128,64) | 1(128) | 0.5 | 2(64,32) | 90% | 80% |
|  | 3(128,64,32) | 2(256,32) | 0.25 | 2(64,32) | 90% | 86% |
|  | 3(128,64,32) | 1(128) | 0.5 | 2(64,32) | **89%** | **87%** |
| Shafatan | 3(128,64,32) | 2(256,32) | 0.2 | - | 80% | 83% |
|  | 3(128,64,32) | 2(256,32) | 0.3 | 2(128,32) | **96%** | **92%** |
|  | baseline architecture [20] |  |  |  | 79% | 75% |
| Thanaya1 | 3(128,64,32) | 2(256,32) | 0.15 | 2(128,32) | **96%** | **90%** |
|  | 3(128,64,32) | 2(256,32) | 0.25 | 2(128,32) | 95% | 89% |
|  | baseline architecture [20] |  |  |  | 79% | 79% |
| Thanaya2 | 3(128,64,32) | 2(256,32) | 0.15 | 2(128,32) | 96% | 90% |
|  | 3(128,64,32) | 2(128,32) | 0.25 | 2(128,32) | **95%** | **90%** |
|  | baseline architecture [20] |  |  |  | 77% | 77% |

Table 3: Group-Specific Classification Models Results

Group-specific classification models results demonstrated that specialized hierarchical models significantly outperform the general-purpose baseline architecture [20] across every single category. For instance, in the Shafatan group, your best configuration reached 92%, significantly outperforming the baseline's 75%. The models showed excellent performance in the Thanaya groups (reaching up to 90% test accuracy), suggesting that partitioning by articulation point successfully reduces confusion between phonetically similar sounds. The performance variance indicates that different phonetic groups require different architectural complexities. For example, groups like Jouf and Shafatan benefited significantly from the inclusion of LSTM layers, whereas some models performed competitively without them, implying that the temporal dependencies of certain articulation points are more easily captured by CNNs and Dense layers alone. Overall, results confirm the effectiveness of the hierarchical approach in successfully reducing class confusion among phonetically similar Arabic letters. While the proposed hierarchical models consistently outperform the baseline architecture [20], we observed a performance gap between training and test sets. As shown in the performance metrics, several models achieved training accuracies exceeding 95%, whereas test accuracies plateaued between 80% and 92%. This discrepancy suggests that the used deep models capacities are too high relative to the size of the Abjad-Kids dataset. For instance, in the Halq group where the training accuracy reached 99%, the model essentially memorized the training samples rather than learning the broader acoustic features of that phonetic class. The dropout values employed (0.1 to 0.5) were essential in curbing extreme divergence; however, the persistent gap indicates that architectural adjustments alone are insufficient. This result reinforces the need for a more expansive dataset.

In addition to alphabet recognition, separate CNN-LSTM models were trained for number and color classification tasks. These tasks involve fewer classes and exhibit lower intra-class similarity compared to alphabet recognition. Colors and numbers classification models are shown in Table 4 and Table 5, respectively. The results indicate that our custom architectures consistently outperform the baseline architecture [3] particularly in the numbers classification task, confirming that deeper feature extraction is advantageous for numeric keyword recognition. • The models that integrated two LSTM layers (e.g., 2(128, 64) or 2(128, 32)) generally maintained more stable test accuracies. This indicates that temporal dependencies are crucial for capturing the distinct phonetic structures of color and number keywords in children's speech. Across both models, there is a consistent generalization gap (ranging from 5% to 14%). For example, in the Colors model, one configuration reached 94% training accuracy but only 82% test accuracy. This suggests that while these models are highly efficient at learning the specific training data distribution, they are reaching a point where additional capacity leads to overfitting rather than better generalization.

| Conv layers | Dens layers | Dropout | LSTM layers | Train Accuracy | Test Accuracy |
|---|---|---|---|---|---|
| 2(128,64) | 3(512,256,32) | 0.3 | 2(128,128) | **93%** | **83%** |
| 2(128,64) | 3(256,64,16) | 0.3 | 2(128,64) | 93% | 79% |
| 3(128,64,16) | 1(128) | 0.5 | 2(64,32) | 94% | 82% |

Table 4: Colors Models Best Results

| Conv layers | Dens layers | Dropout | LSTM layers | Train Accuracy | Test Accuracy |
|---|---|---|---|---|---|
| 3(128,64,32) | 3(512,256,32) | 0.3 | 2(128,128) | 95% | 83% |
| 3(128,64,32) | 3(256,128,16) | 0.3 | 2(128,64) | **91%** | **86%** |
| baseline architecture [20] | | | | 89% | 80% |

Table 5: Number Models Best Results

As for the effect of data augmentation, data augmentation played a critical role in improving model generalization. Without augmentation, models exhibited overfitting. Applying pitch shifting, low-pass filtering, and gain adjustment increased data diversity and improved test accuracy across all tasks. Augmentation was particularly effective for alphabet groups with fewer samples per class.

6. Conclusion

Speech-based AI has the power to transform how children learn, and large pretrained and multiscale transformer systems show strong performance on general audio benchmarks. However, low-resource languages like Arabic suffer from data scarcity. Moreover, the lack of high-quality child-specific datasets is a critical gap that poses essential challenges. To bridge this gap, we introduce Abjad-Kids, an Arabic speech dataset for kindergarten and primary education. Covering 141 classes including alphabets, numbers, and colors. Our dataset contains 46,397 audio samples from children aged 3 to 12. Recognizing such great number of classes from children speech is an acoustically complex problem, especially that Arabic phonemes often overlap. Thus, we developed a hierarchical two-stage classification methodology using CNN-LSTM architectures. First stage identifies the phonetic group of a word, and the second stage performs the alphabet classes classification. Our experiments show that grouping sounds based on how they are articulated (static grouping) provides a more robust foundation than purely data-driven methods. While our models consistently outperform an existing baseline, they also highlight the persistent challenge of overfitting. We view this limitation as an invitation for the research community to expand our dataset. We hope that Abjad-Kids enrich the children's representation in speech dataset and would be a step toward more inclusive, responsive, and effective AI tools that can truly support Arabic-speaking children in their early learning journey.

## 7. Limitations and Future Work

Despite achieving promising results, most of our experiments indicate a challenge with overfitting. This is likely due to data scarcity, as we still have limited number of samples available per class, even after applying data augmentation techniques. While regularization methods, such as dropout and weight decay, were employed to mitigate this, they provided only marginal improvements in generalization performance.

Future research will prioritize the expansion of the Abjad-Kids dataset to increase class-level diversity. Furthermore, we intend to explore transfer learning and more advanced synthetic data generation techniques to address the imbalance between model capacity and available training data. Moreover, future work may explore advanced architectures such as lightweight transformers or self-supervised audio representation learning to enhance performance under limited data conditions.

**Funding**

The authors declare that they have no funding.